\documentclass[10pt,english,oneside,twocolumn,a4paper]{article}
\usepackage{graphicx}
\usepackage[utf8]{inputenc}
\usepackage{times}
\fontfamily{ptm}\selectfont
\usepackage{tabularx}
\usepackage{ragged2e}
\usepackage[singlelinecheck=false]{caption}
\usepackage{subcaption}
\usepackage[T1]{fontenc}
\usepackage{csquotes}
\usepackage{amsmath}
\usepackage{amssymb}
\usepackage{graphicx,booktabs}
\usepackage[table]{xcolor}
\usepackage[
  hidelinks,
  pdftitle={Prompting Foundation Models for Zero-Shot Ship Instance Segmentation in SAR Imagery},
  pdfauthor={Islam Mansour, Francescopaolo Sica, Michael Schmitt},
  pdfsubject={SAR Image Analysis, Instance Segmentation, Foundation Models},
  pdfkeywords={SAR, ship detection, instance segmentation, SAM2, YOLOv11, zero-shot, maritime surveillance},
  pdfcreator={LaTeX with hyperref},
  pdfproducer={pdfLaTeX},
]{hyperref}
\usepackage{float}

\usepackage{orcidlink}
\usepackage[style=ieee,doi=false,isbn=false,url=false,eprint=false,date=year]{biblatex}
\AtEveryBibitem{%
  \clearname{editor}
  \clearfield{venue}
  \clearfield{series}
  \clearlist{language} 
  \clearfield{pages}
  \ifentrytype{online}{}{%
    \clearfield{url}
  }

  \iffieldundef{title}{}{
    \savefield{title}{\tmpfield}  
    \def\formattedtitle{{\tmpfield}}  
    \restorefield{title}{\formattedtitle}  
  }
}

\DeclareFieldFormat{eid}{art no.\addabbrvspace #1} 

\DefineBibliographyStrings{english}{
  references = {Literature}
}

\usepackage{babel}
\RequirePackage[blocks]{authblk}
\makeatletter
\let\ps@plain\ps@empty
\def\@xivpt{14pt}
\def\@sect#1#2#3#4#5#6[#7]#8{%
  \ifnum #2<2
    \null\par\vskip-15pt
  \fi
  \ifnum #2>\c@secnumdepth 
    \let\@svsec\@empty
  \else
    \refstepcounter{#1}%
    \protected@edef\@svsec{%
      \ifnum #2<4
        \hb@xt@10mm{\csname the#1\endcsname}\relax
      \else
        \hb@xt@12mm{\csname the#1\endcsname}\relax
      \fi}%
  \fi
  \@tempskipa #5\relax
  \ifdim \@tempskipa>\z@
    \begingroup
      #6{%
        \@hangfrom{\hskip #3\relax\@svsec}%
          \interlinepenalty \@M #8\@@par}%
    \endgroup
    \csname #1mark\endcsname{#7}%
    \addcontentsline{toc}{#1}{%
      \ifnum #2>\c@secnumdepth \else  
        \protect\numberline{\csname the#1\endcsname}%
      \fi 
      #7}%
  \else
    \def\@svsechd{%
      #6{\hskip #3\relax
      \@svsec #8}%
      \csname #1mark\endcsname{#7}%
      \addcontentsline{toc}{#1}{%
        \ifnum #2>\c@secnumdepth \else
          \protect\numberline{\csname the#1\endcsname}%
        \fi
        #7}}%
  \fi
  \@xsect{#5}}
\renewcommand\LARGE{\@setfontsize\LARGE{16}{20}}
\def\abstract#1{\def\@abstract{#1}}
\def\abstractEn#1{\def\@abstractEn{#1}}
\def\titleEn#1{\def\@titleEn{#1}}
\def\@maketitle{%
  \newpage
  \null
  \let \footnote \thanks
    {\LARGE\bfseries\RaggedRight \@title \par}%
    \vskip 1\baselineskip%
    {\normalsize
      \@author\par}%
    \vskip 2\baselineskip%
    \vskip \baselineskip%
    {\section*{Abstract}
      \@abstract}%
  \par
  \vskip 3\baselineskip}

\renewcommand\section{\@startsection {section}{1}{\z@}%
                                   {-3.5ex \@plus -1ex \@minus -.2ex}%
                                   {\baselineskip}%
                                   {\normalfont\Large\bfseries\RaggedRight}}
\renewcommand\subsection{\@startsection{subsection}{2}{\z@}%
                                     {\baselineskip}%
                                     {1ex}%
                                     {\normalfont\large\bfseries\RaggedRight}}
\renewcommand\subsubsection{\@startsection{subsubsection}{3}{\z@}%
                                     {1\baselineskip}%
                                     {3bp}%
                                     {\normalfont\normalsize\bfseries\RaggedRight}}
\renewcommand\paragraph{\@startsection{paragraph}{4}{\z@}%
                                    {1\baselineskip\@plus1ex \@minus.2ex}%
                                    {3bp}%
                                    {\normalfont\normalsize\RaggedRight}}
\renewcommand\subparagraph{\@startsection{subparagraph}{5}{\parindent}%
                                       {3.25ex \@plus1ex \@minus .2ex}%
                                       {-1em}%
                                      {\normalfont\normalsize\bfseries\RaggedRight}}
\parindent\p@
\parskip\z@
\DeclareCaptionLabelSeparator{enskip}{\enskip}
\makeatother

\title{Prompting Foundation Models for Zero-Shot Ship Instance Segmentation in SAR Imagery}

\author[a]{Islam Mansour\,\orcidlink{0000-0003-3114-6515}}
\author[a]{Francescopaolo Sica\,\orcidlink{0000-0003-1593-1492}}
\author[a]{Michael Schmitt\,\orcidlink{0000-0002-0575-2362}}
\affil[a]{Department of Aerospace Engineering, University of the Bundeswehr Munich, Neubiberg, Germany}

\abstract{Synthetic Aperture Radar (SAR) plays a critical role in maritime surveillance, yet deep learning for SAR analysis is limited by the lack of pixel-level annotations. This paper explores how general-purpose vision foundation models can enable zero-shot ship instance segmentation in SAR imagery, eliminating the need for pixel-level supervision. A YOLOv11-based detector trained on open SAR datasets localizes ships via bounding boxes, which then prompt the Segment Anything Model 2 (SAM2) to produce instance masks without any mask annotations. 
Unlike prior SAM-based SAR approaches that rely on fine tuning or adapters, our method demonstrates that spatial constraints from a SAR-trained detector alone can effectively regularize foundation model predictions.
This design partially mitigates the optical--SAR domain gap and enables downstream applications such as vessel classification, size estimation, and wake analysis.
Experiments on the SSDD benchmark achieve a mean IoU of~0.637 (89\% of a fully supervised baseline) with an overall ship detection rate of 89.2\%, confirming a scalable, annotation-efficient pathway toward foundation-model-driven SAR image understanding.}

\begin{document}

\maketitle

\section{Introduction}

Maritime surveillance plays a critical role in global trade, security, and environmental management. Synthetic Aperture Radar (SAR) has become central to this domain: as an active microwave sensor it offers day-and-night, all-weather operation and broad-area coverage~\cite{sorensen_maritime_2024,wang_automatic_2019,wang_data-driven_2023}.

Traditional SAR analysis has primarily focused on bounding box detection. While effective for localization, boxes offer limited semantic detail for vessel classification or anomaly detection~\cite{li_ship_2017}. Instance segmentation provides pixel-level masks enabling precise estimates of vessel size, orientation, and structure, but typically requires fully supervised models trained on dense pixel-level annotations~\cite{qiao_review_2025,zhang_ls-ssdd-v10_2020}.

In SAR, generating such labels is particularly challenging due to backscatter, speckle, and imaging artifacts—requiring expert knowledge that makes large-scale annotation impractical~\cite{zhang_sar_2021}. Bounding box annotations are far easier to collect, motivating methods that minimize supervision without sacrificing segmentation quality.

This work proposes an annotation-efficient framework using only bounding box supervision: a YOLOv11-based detector localizes ships, and these boxes prompt SAM2 to produce zero-shot pixel-level masks. Our contributions are: (1) a zero-shot SAR segmentation pipeline via detection-driven prompting; (2) an analysis of how spatial constraints from a domain-specific detector mitigate the SAR–optical modality gap; and (3) a quantitative comparison with supervised baselines and an inference cost analysis.

\section{Related Work}

Recent advances in vision foundation models (FMs) and SAR-specific detection have opened new pathways for zero-shot segmentation in remote sensing. The Segment Anything Model (SAM)~\cite{kirillov_segment_2023}, trained on over one billion masks, introduced a promptable segmentation framework for zero-shot inference. Its successor, SAM2~\cite{ravi_sam_2024}, adds a hierarchical vision transformer (Hiera) backbone providing multi-scale spatial features—beneficial for small SAR targets—with improved mask quality and faster inference. However, transfer to SAR remains limited due to speckle noise, radar-specific backscatter, and geometric distortions absent from optical training data~\cite{inkawhich_status_2025}.

Several recent works have attempted to bridge this domain gap by adapting SAM to SAR imagery. SAMSAR~\cite{rahimi_samsar_2025} fine-tunes SAM with adapter modules on SAR data, while Zheng et al.~\cite{zheng_tuning_2025} explore parameter-efficient tuning strategies, and Liu et al.~\cite{liu_context-aggregated_2024} incorporate context aggregation into the prompt encoder. While effective, these approaches all require SAR-specific mask annotations or architectural modifications—reintroducing the supervision burden that foundation models aim to eliminate. In contrast, our approach uses SAM2 entirely off the shelf, relying solely on bounding box prompts derived from a SAR-trained detector to achieve zero-shot segmentation without any fine-tuning, adapter modules, or pixel-level SAR labels.

YOLO-based detectors have proven effective for SAR ship detection due to their speed and robustness to small targets and clutter. Variants such as BiFA-YOLO~\cite{sun_bifa-yolo_2021} and contextual region-based CNNs~\cite{kang_contextual_2017} achieve strong performance on SSDD~\cite{zhang_sar_2021} using only bounding box annotations.

Prompt-based pipelines that decouple detection and segmentation have emerged as a flexible solution across computer vision, medical imaging, and remote sensing~\cite{cheng_sam_2023,ma_segment_2024}. This detect-then-prompt architecture is well-suited to SAR, where annotation costs are high and cross-domain generalization is nontrivial. Our work adopts this paradigm, using SAR-trained bounding boxes from YOLOv11 as spatial prompts for zero-shot segmentation with SAM2.

\section{Methodology}

We propose a two-stage detect-and-prompt pipeline (Figure~\ref{fig:workflow}). Stage 1: a YOLOv11-based detector trained on bounding box-labeled SAR data localizes ship instances. Stage 2: each detected box $B_i$ serves as a spatial prompt for SAM2, which generates pixel-level instance masks $M_i = \mathrm{SAM2}(I, B_i)$ in a zero-shot manner—without fine-tuning or SAR-specific adaptation.

This decoupled design lets each component focus on its strengths: YOLOv11 handles domain-specific detection under speckle and clutter, while SAM2 leverages its generalization capacity for segmentation. Bounding box prompts mitigate the optical–SAR modality gap by constraining SAM2 to localized ship regions, enabling vessel classification, size estimation, and wake analysis without any mask supervision.

Each detected bounding box is passed to SAM2 as a four-coordinate box prompt. SAM2 returns up to three candidate masks per prompt ranked by predicted IoU; we select the highest-scoring mask and apply a confidence threshold of 0.5 on the YOLO detections to filter low-quality prompts. No additional point prompts or post-processing are used.

The source code is publicly available at \url{https://github.com/IslamAlam/hybrivision}.

\begin{figure}[htbp]
\centering
\includegraphics[width=\linewidth]{figures/workflow.drawio.png}
\caption{Two-stage detect-and-prompt framework. Stage 1: YOLOv11 detects ship instances and generates bounding box prompts. Stage 2: SAM2 produces zero-shot segmentation masks from these prompts.}
\label{fig:workflow}
\end{figure}

\section{Experiments}

\subsection{Data}
\begin{figure*}[!t]
\centering
\includegraphics[width=0.92\textwidth]{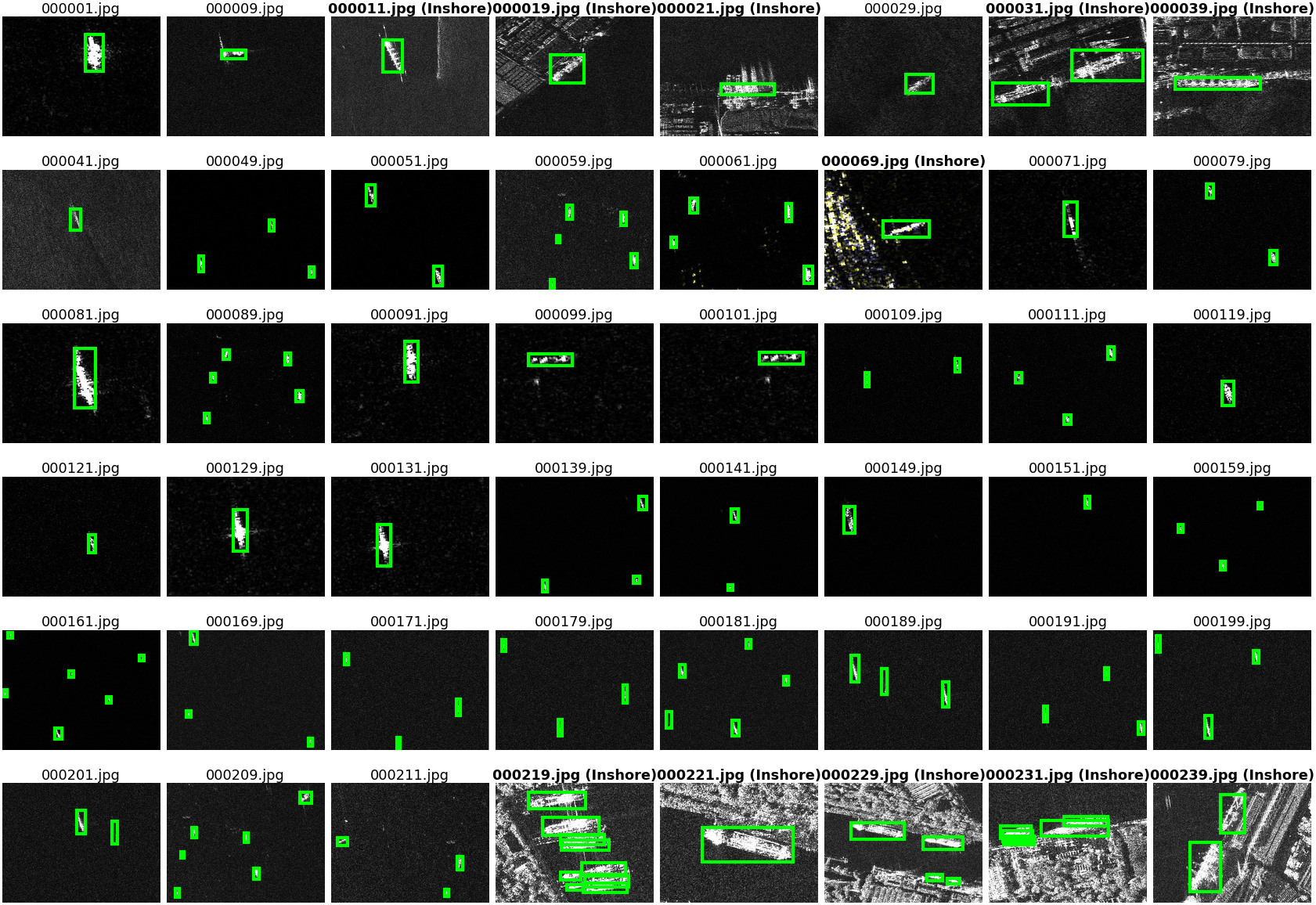}
\caption{HBB ground truths for 48 SSDD test images (000001--000239, 6$\times$8 grid). Inshore samples are labelled accordingly.}
\label{fig:dataset_hbb}
\end{figure*}

\begin{figure*}[!t]
\centering
\includegraphics[width=0.92\textwidth]{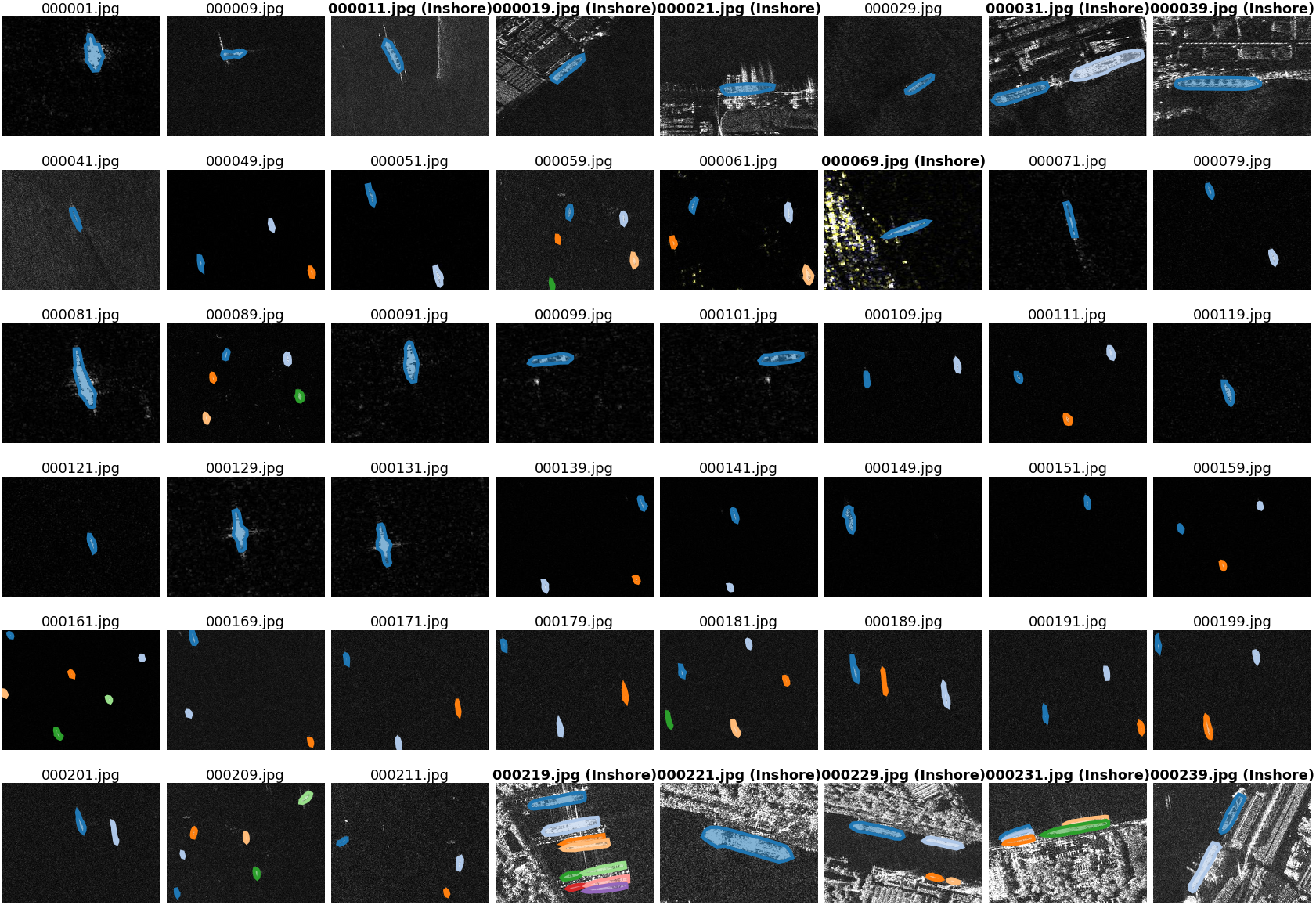}
\caption{Polygon segmentation ground truths for 48 SSDD test images (000001--000239, 6$\times$8 grid). Each instance is shown in a distinct colour. Inshore samples are labelled accordingly.}
\label{fig:dataset_pseg}
\end{figure*}

All experiments use the SAR Ship Detection Dataset (SSDD)~\cite{zhang_sar_2021}: 1,160 images from Sentinel-1, RadarSat-2, and TerraSAR-X at 1--15\,m/px resolution, with 2,358 ship instances annotated with bounding boxes and polygon masks across inshore and offshore subsets. Figures~\ref{fig:dataset_hbb}--\ref{fig:dataset_pseg} show HBB and polygon ground truths for 48 test images, with inshore samples labelled accordingly.

While SSDD is the most widely adopted benchmark for SAR ship segmentation and enables direct comparison with prior work, it is limited in geographic diversity and sensor coverage. Validating the proposed pipeline on additional datasets such as HRSID~\cite{wei_hrsid_2020} or LS-SSDD-v1.0~\cite{zhang_ls-ssdd-v10_2020} remains an important direction for assessing cross-dataset generalization, which we leave to future work.

\begin{figure*}[!htbp]
    \centering
    \includegraphics[width=0.85\textwidth]{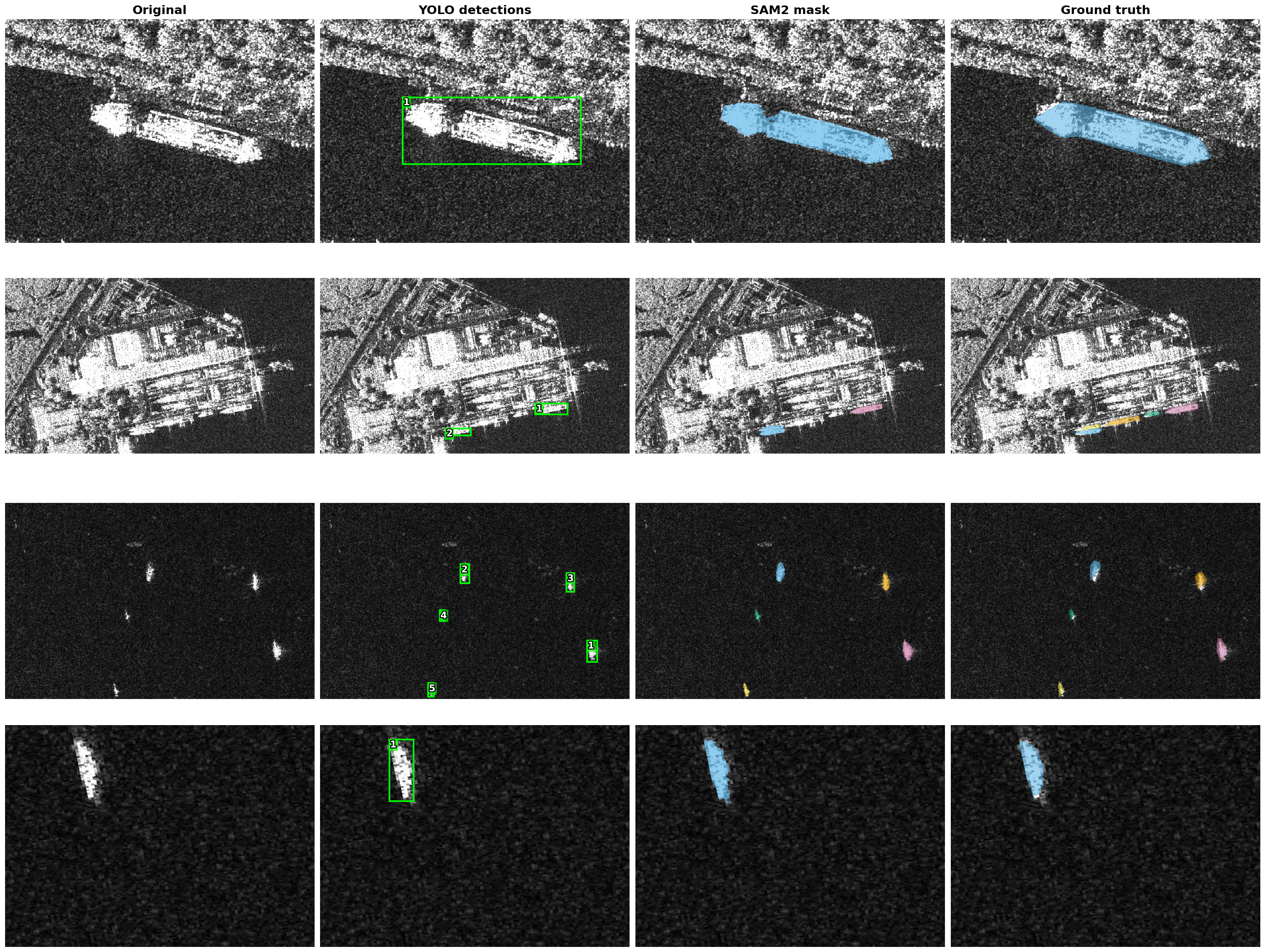}
    \caption{Qualitative segmentation results of the YOLOv11 + SAM2 pipeline on the SSDD test set. Columns: original SAR image, YOLOv11 bounding box detections, SAM2-predicted instance masks, ground truth polygon masks. Top rows show inshore scenes; bottom rows show offshore scenes. In dense inshore scenes SAM2 occasionally merges adjacent vessels.}
    \label{fig:yolo_sam2_seg_results}
\end{figure*}

\begin{figure*}[!htbp]
    \centering
    \includegraphics[width=0.85\textwidth]{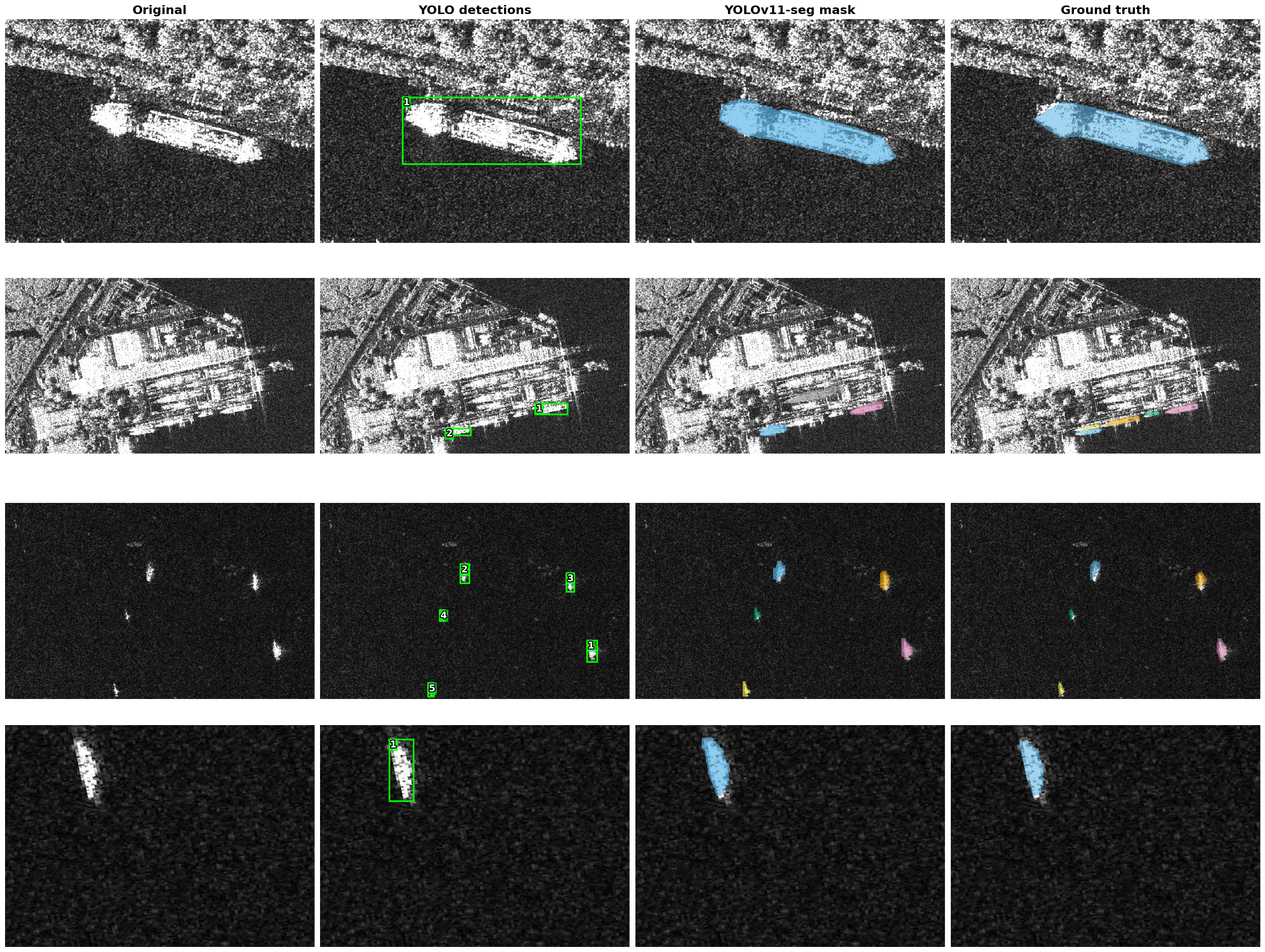}
    \caption{Qualitative segmentation results of the supervised YOLOv11-seg baseline on the same SSDD test images. Columns: original SAR image, YOLO detections, YOLOv11-seg predicted masks, ground truth polygon masks.}
    \label{fig:yolo_seg_results}
\end{figure*}

\subsection{Evaluation Protocol}

The detector is trained and validated using mean Average Precision (mAP) following the COCO protocol:
\begin{equation}
    \mathrm{mAP} = \frac{1}{|T|} \sum_{t \in T} \mathrm{AP}_{\mathrm{IoU}=t},
\end{equation}
where $T$ is the set of IoU thresholds from $0.5$ to $0.95$ in increments of $0.05$.

Segmentation quality is evaluated by comparing SAM2-predicted masks against ground-truth polygon annotations using Intersection over Union (IoU) and the Dice coefficient (F1 score):
\begin{equation}
    \mathrm{IoU} = \frac{|M_{\mathrm{pred}} \cap M_{\mathrm{gt}}|}{|M_{\mathrm{pred}} \cup M_{\mathrm{gt}}|},
\end{equation}
\begin{equation}
    \mathrm{Dice} = \frac{2\,|M_{\mathrm{pred}} \cap M_{\mathrm{gt}}|}{|M_{\mathrm{pred}}| + |M_{\mathrm{gt}}|},
\end{equation}
where $M_{\mathrm{pred}}$ and $M_{\mathrm{gt}}$ denote the predicted and ground-truth binary masks, respectively. We additionally report pixel-level precision and recall, as well as a relaxed IoU computed after morphological dilation of both masks to account for annotation boundary noise.

\subsection{Implementation}

YOLOv11 is trained with the AdamW optimizer and cosine learning rate scheduling in PyTorch. SAM2 is used directly from Meta AI's public release with the Hiera-L backbone, without any fine-tuning or domain adaptation.

\subsection{Results and Analysis}

\begin{table*}[htbp]
\footnotesize
\caption{Segmentation performance of the YOLOv11 + SAM2 framework on the SSDD test set (N=232). Rel.~IoU: relaxed IoU with 1-pixel boundary dilation. Det.~Rate: YOLO per-instance box recall (IoU$_\text{match}$\,=\,0.5).}
\centering
\begin{tabular}{lccccccccccc}
\toprule
\textbf{Scene} & \textbf{N} & \textbf{IoU mean} & \textbf{IoU std} & \textbf{IoU med} & \textbf{Dice/F1} & \textbf{Precision} & \textbf{Recall} & \textbf{IoU@0.50} & \textbf{IoU@0.75} & \textbf{Rel.~IoU} & \textbf{Det.~Rate} \\
\midrule
Inshore  & 46  & 0.571 & 0.200 & 0.637 & 0.702 & 0.676 & 0.820 & 0.674 & 0.152 & 0.598 & 0.878 \\
Offshore & 186 & 0.653 & 0.147 & 0.678 & 0.779 & 0.774 & 0.823 & 0.871 & 0.301 & 0.680 & 0.920 \\
\addlinespace
Overall  & 232 & 0.637 & 0.162 & 0.667 & 0.764 & 0.755 & 0.822 & 0.832 & 0.272 & 0.664 & 0.892 \\
\bottomrule
\end{tabular}
\label{tab:results}
\end{table*}

Table~\ref{tab:results} shows that the pipeline achieves an overall detection rate of 89.2\% (YOLO box recall) and a mean IoU of 0.637 (median 0.667, Dice 0.764) across 232 SSDD test images. Detection rate is higher in offshore scenes (92.0\%) than in inshore ones (87.8\%), consistent with the greater clutter and dense ship packing in the latter. Offshore scenes also outperform inshore in segmentation quality (IoU 0.653 vs.\ 0.571). Notably, mask recall is nearly identical across both settings (0.820 vs.\ 0.823), indicating that SAM2 captures a similar fraction of ship pixels regardless of scene complexity; the segmentation quality gap is instead driven by lower precision in inshore scenes (0.676 vs.\ 0.774), where SAM2 masks bleed into adjacent structures and neighboring vessels. The IoU@0.75 gap (0.301 vs.\ 0.152) further highlights boundary precision as the primary challenge in dense inshore scenes. Relaxed IoU improves from 0.637 to 0.664 at $r=1$\,px dilation, indicating that SAM2 predictions are systematically sharper than the coarse ground-truth polygons—demonstrating that the model recovers fine ship geometry despite lacking SAR-specific training.

\begin{figure}[!htbp]
\centering
\begin{subfigure}[b]{0.48\textwidth}
    \centering
    \includegraphics[width=\linewidth]{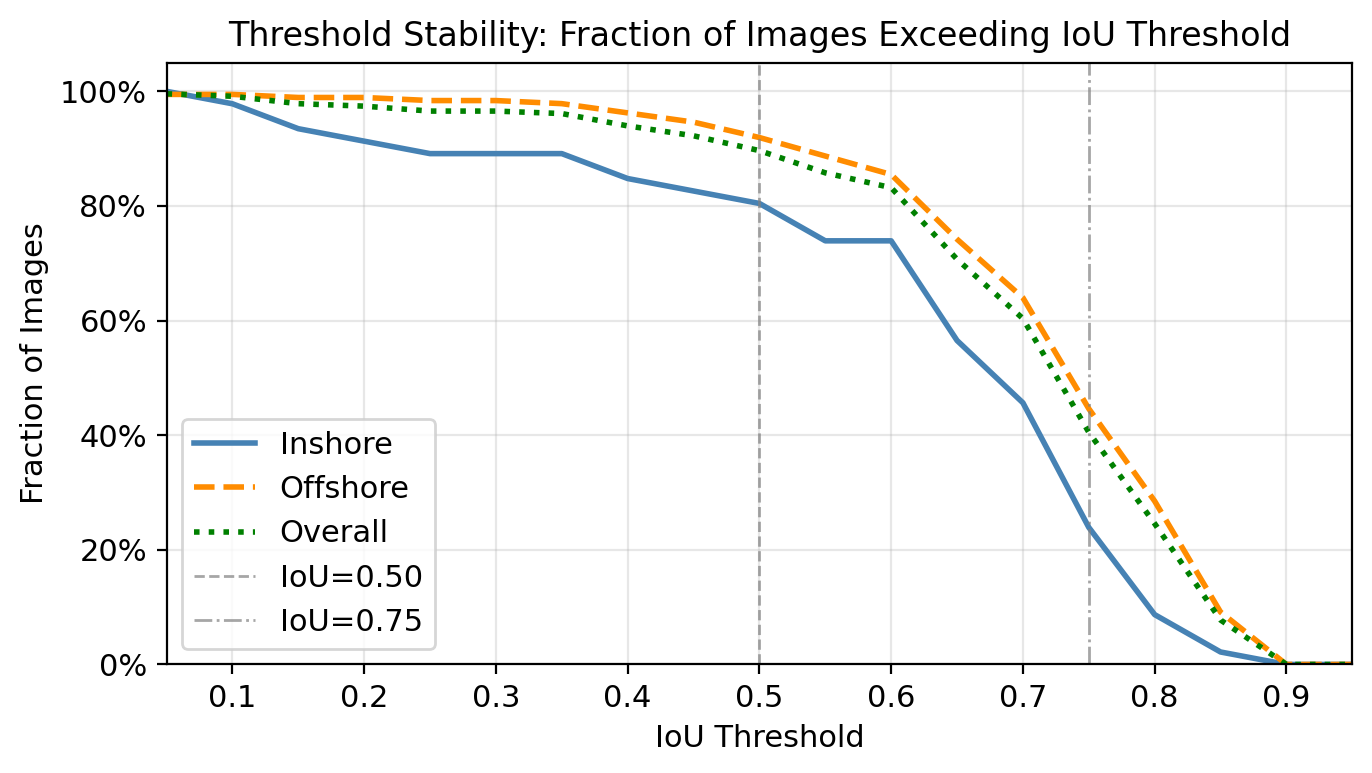}
    \caption{Fraction of instances whose IoU meets or exceeds a given threshold.}
    \label{fig:threshold}
\end{subfigure}
\vfill
\begin{subfigure}[b]{0.48\textwidth}
    \centering
    \includegraphics[width=\linewidth]{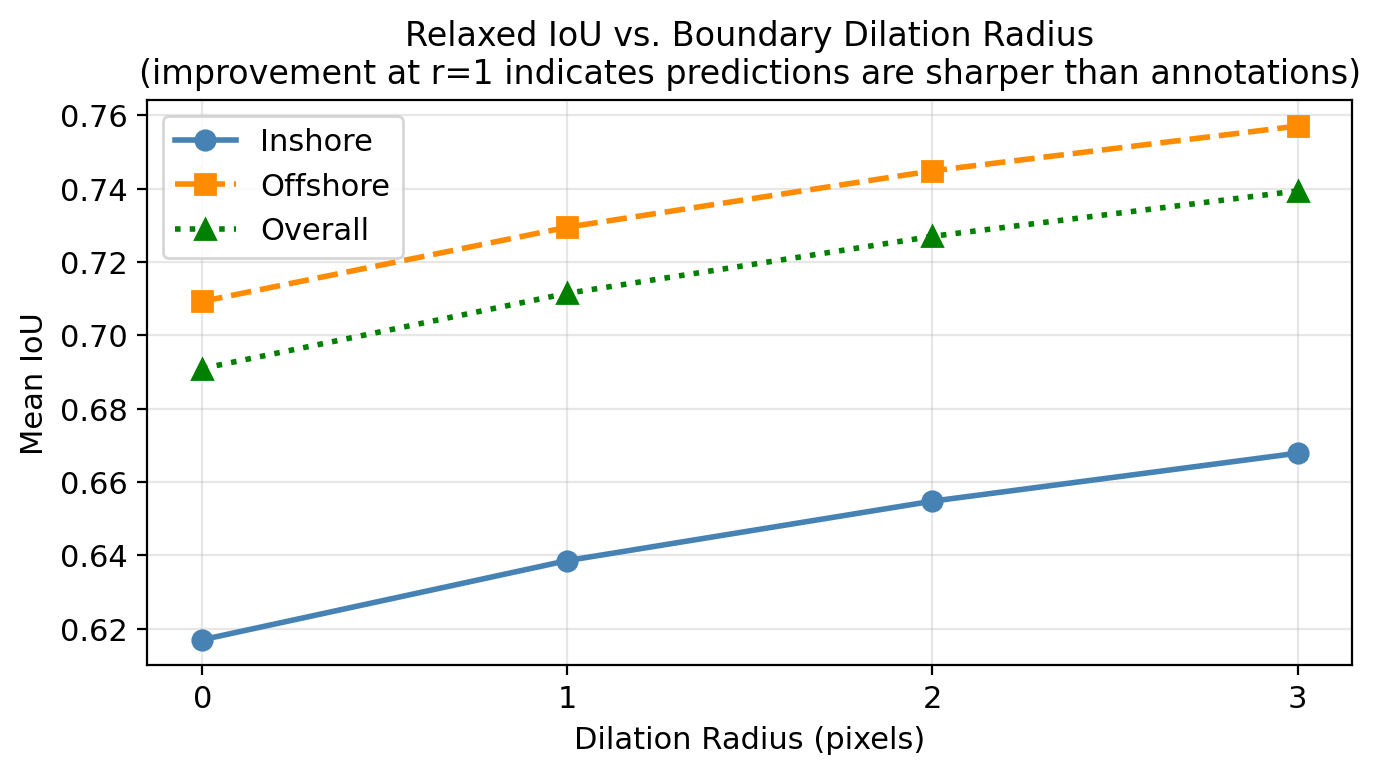}
    \caption{Relaxed IoU vs.\ boundary dilation radius.}
    \label{fig:relaxed}
\end{subfigure}
\caption{Performance analysis: (a) Offshore predictions remain robust up to IoU~$=$~0.65; inshore degrades more steeply due to background clutter and merged vessels. (b) IoU improves at $r=1$~px across all subsets, indicating SAM2 predictions are systematically sharper than the coarse ground-truth annotations.}
\label{fig:performance_analysis}
\end{figure}

\subsubsection*{Qualitative results}
Figures~\ref{fig:yolo_sam2_seg_results} and~\ref{fig:yolo_seg_results} compare zero-shot SAM2 masks against the supervised baseline across small vessels, cluttered ports, and varied orientations. Failure modes include merged masks for closely packed ships and occasional over-segmentation from inaccurate prompts. Adding positive and negative point prompts within and outside each detection box has been shown to improve boundary recognition in related domains~\cite{jiangPromptEngineeringSegment2025} and could resolve these inshore failures.

Figure~\ref{fig:threshold} shows the fraction of instances exceeding each IoU threshold: the offshore curve remains above 80\% up to IoU~$=$~0.65, while the inshore curve drops more steeply. Figure~\ref{fig:relaxed} confirms that relaxed IoU improves at small radii ($r\le1$\,px) before saturating, further evidence that predicted boundaries are tighter than ground-truth polygons.

\subsubsection*{Comparison with supervised baseline}
Table~\ref{tab:comparison} compares our pipeline against YOLOv11-seg—the same backbone with a mask prediction head trained on SSDD polygon annotations. Our zero-shot approach achieves approximately 89\% of the supervised IoU on offshore scenes with no pixel-level annotation; the remaining gap is concentrated in inshore scenes where densely packed ships are difficult to separate without negative prompts.

\begin{table}[htbp]
\caption{Comparison with the supervised YOLOv11-seg baseline on the SSDD test set.}
\centering
\footnotesize
\begin{tabular}{llccc}
\toprule
\textbf{Method} & \textbf{Supervision} & \textbf{Scene} & \textbf{IoU} & \textbf{Dice} \\
\midrule
YOLO+SAM2 (ours) & bbox only & Inshore  & 0.571 & 0.702 \\
YOLO+SAM2 (ours) & bbox only & Offshore & 0.653 & 0.779 \\
YOLO+SAM2 (ours) & bbox only & Overall  & 0.637 & 0.764 \\
\addlinespace
YOLOv11-seg      & masks     & Inshore  & 0.615 & 0.736 \\
YOLOv11-seg      & masks     & Offshore & 0.736 & 0.843 \\
YOLOv11-seg      & masks     & Overall  & 0.712 & 0.822 \\
\bottomrule
\end{tabular}
\label{tab:comparison}
\end{table}

\begin{table}[htbp]
\caption{Inference time breakdown on an NVIDIA GeForce RTX 3090 GPU (mean $\pm$ std over 20 images).}
\centering
\footnotesize
\begin{tabular}{lcc}
\toprule
\textbf{Stage} & \textbf{Time (ms/image)} & \textbf{Notes} \\
\midrule
YOLOv11 detection  & $8 \pm 2$    & constant per image \\
SAM2 segmentation  & $85 \pm 12$  & per detected ship \\
Total (offshore)   & $\approx 178$ & $\sim$2 ships/image \\
Total (inshore)    & $\approx 348$ & $\sim$4 ships/image \\
\bottomrule
\end{tabular}
\label{tab:timing}
\end{table}

\subsubsection*{Computational cost}
SAM2 (Hiera-L) dominates latency at $\approx$\,85\,ms/ship vs.\ YOLO's $\approx$\,8\,ms/image, scaling linearly with density: $\approx$\,178\,ms for offshore ($\sim$2 ships) and $\approx$\,348\,ms for inshore ($\sim$4 ships) on an NVIDIA GeForce RTX 3090 GPU (Table~\ref{tab:timing}). This overhead is modest relative to satellite revisit intervals and can be reduced with smaller SAM2 backbones (e.g.\ Hiera-T at $\approx$\,28\,ms per ship) or per-image prompt batching.

\subsubsection*{Limitations}
The primary limitation of the proposed pipeline is the merging of masks for closely spaced ships in dense inshore scenes, stemming from the use of bounding box prompts alone without negative point cues. Additionally, the evaluation is conducted on a single benchmark (SSDD); while this enables comparison with prior work, cross-dataset validation on HRSID or LS-SSDD-v1.0 is needed to establish generalization across sensors and geographic regions. Finally, the current evaluation uses a single SAM2 configuration (Hiera-L); a systematic study of backbone size, prompt type, and detection quality on segmentation performance would further characterize the pipeline's robustness.

\section{Conclusion}

We present a zero-shot ship instance segmentation pipeline for SAR that combines a domain-specific YOLOv11 detector with SAM2. Using only bounding box annotations, the pipeline achieves 0.637 mean IoU on SSDD—89\% of the fully supervised YOLOv11-seg baseline—with an overall detection rate of 89.2\% (87.8\% inshore, 92.0\% offshore), while producing masks geometrically finer than polygon ground truths.

Future work will explore multi-point SAM2 prompting with positive and negative cues to improve instance separation in dense scenes~\cite{jiangPromptEngineeringSegment2025}, few-shot SAM2 fine-tuning on a small SAR mask set, cross-dataset evaluation on HRSID and LS-SSDD-v1.0, and unified detection–segmentation architectures to close the remaining gap in challenging inshore scenarios.

\printbibliography

@article{jiangPromptEngineeringSegment2025,
	title        = {Prompt Engineering in Segment Anything Model: Methodologies, Applications, and Emerging Challenges},
	shorttitle   = {Prompt Engineering in Segment Anything Model},
	author       = {Jiang, Yidong},
	year         = 2025,
	journal      = {Corr},
	volume       = {abs/2507.9562},
	doi          = {10.48550/ARXIV.2507.09562},
	urldate      = {2025-11-06},
	eprint       = {2507.09562},
	archiveprefix = {arXiv},
}

@article{cheng_sam_2023,
	title        = {{SAM} on Medical Images: {A} Comprehensive Study on Three Prompt Modes},
	author       = {Dongjie Cheng and Ziyuan Qin and Zekun Jiang and Shaoting Zhang and Qicheng Lao and Kang Li},
	year         = 2023,
	journal      = {CoRR},
	volume       = {abs/2305.00035},
	doi          = {10.48550/ARXIV.2305.00035},
	url          = {https://doi.org/10.48550/arXiv.2305.00035},
	eprinttype   = {arXiv},
	eprint       = {2305.00035},
	bibsource    = {dblp computer science bibliography, https://dblp.org}
}

@article{ma_segment_2024,
	title        = {Segment anything in medical images},
	author       = {Ma, Jun and He, Yuting and Li, Feifei and Han, Lin and You, Chenyu and Wang, Bo},
	year         = 2024,
	journal      = {Nature Communications},
	publisher    = {Nature Publishing Group UK London},
	volume       = 15,
	number       = 1,
	pages        = 654
}

@article{zheng_tuning_2025,
	title        = {Tuning a SAM-Based Model With Multicognitive Visual Adapter to Remote Sensing Instance Segmentation},
	author       = {Linghao Zheng and Xinyang Pu and Su Zhang and Feng Xu},
	year         = 2025,
	journal      = {{IEEE} J. Sel. Top. Appl. Earth Obs. Remote. Sens.},
	volume       = 18,
	pages        = {2737--2748},
	doi          = {10.1109/JSTARS.2024.3504409},
	url          = {https://doi.org/10.1109/JSTARS.2024.3504409},
	bibsource    = {dblp computer science bibliography, https://dblp.org}
}

@article{rahimi_samsar_2025,
	title        = {{SAMSAR:} {A} modified {SAM} architecture for oceanic ship segmentation of satellite {SAR} images using CNN-based Cross-Fused Attention},
	author       = {Mahdi Rahimi and Saeed Sharifian},
	year         = 2025,
	journal      = {Expert Syst. Appl.},
	volume       = 284,
	pages        = 127852,
	doi          = {10.1016/J.ESWA.2025.127852},
	url          = {https://doi.org/10.1016/j.eswa.2025.127852},
	bibsource    = {dblp computer science bibliography, https://dblp.org}
}

@article{liu_context-aggregated_2024,
	title        = {Context-Aggregated and SAM-Guided Network for ViT-Based Instance Segmentation in Remote Sensing Images},
	author       = {Shuangzhou Liu and Feng Wang and Hongjian You and Niangang Jiao and Guangyao Zhou and Tingtao Zhang},
	year         = 2024,
	journal      = {Remote. Sens.},
	volume       = 16,
	number       = 13,
	pages        = 2472,
	doi          = {10.3390/RS16132472},
	url          = {https://doi.org/10.3390/rs16132472},
	bibsource    = {dblp computer science bibliography, https://dblp.org}
}

@article{qiao_review_2025,
	title        = {A Review of Deep-Learning-Based {SAR} Image Ship Interpretation Technology: The Latest Advances},
	author       = {Sijia Qiao and Qingjun Zhang and Zhibin Wang},
	year         = 2025,
	journal      = {{IEEE} J. Sel. Top. Appl. Earth Obs. Remote. Sens.},
	volume       = 18,
	pages        = {26152--26185},
	doi          = {10.1109/JSTARS.2025.3614504},
	url          = {https://doi.org/10.1109/JSTARS.2025.3614504},
	bibsource    = {dblp computer science bibliography, https://dblp.org}
}

@article{inkawhich_status_2025,
	title        = {On the Status of Foundation Models for {SAR} Imagery},
	author       = {Nathan Inkawhich},
	year         = 2025,
	journal      = {CoRR},
	volume       = {abs/2509.21722},
	doi          = {10.48550/ARXIV.2509.21722},
	url          = {https://doi.org/10.48550/arXiv.2509.21722},
	eprinttype   = {arXiv},
	eprint       = {2509.21722},
	bibsource    = {dblp computer science bibliography, https://dblp.org}
}

@article{ravi_sam_2024,
	title        = {{SAM} 2: Segment Anything in Images and Videos},
	author       = {Nikhila Ravi and Valentin Gabeur and Yuan{-}Ting Hu and Ronghang Hu and Chaitanya Ryali and Tengyu Ma and Haitham Khedr and Roman R{\"{a}}dle and Chlo{\'{e}} Rolland and Laura Gustafson and Eric Mintun and Junting Pan and Kalyan Vasudev Alwala and Nicolas Carion and Chao{-}Yuan Wu and Ross B. Girshick and Piotr Doll{\'{a}}r and Christoph Feichtenhofer},
	year         = 2024,
	journal      = {CoRR},
	volume       = {abs/2408.00714},
	doi          = {10.48550/ARXIV.2408.00714},
	url          = {https://doi.org/10.48550/arXiv.2408.00714},
	eprinttype   = {arXiv},
	eprint       = {2408.00714},
	bibsource    = {dblp computer science bibliography, https://dblp.org}
}

@inproceedings{kirillov_segment_2023,
	title        = {Segment Anything},
	author       = {Alexander Kirillov and Eric Mintun and Nikhila Ravi and Hanzi Mao and Chlo{\'{e}} Rolland and Laura Gustafson and Tete Xiao and Spencer Whitehead and Alexander C. Berg and Wan{-}Yen Lo and Piotr Doll{\'{a}}r and Ross B. Girshick},
	year         = 2023,
	booktitle    = {{IEEE/CVF} International Conference on Computer Vision, {ICCV} 2023, Paris, France, October 1-6, 2023},
	publisher    = {{IEEE}},
	pages        = {3992--4003},
	doi          = {10.1109/ICCV51070.2023.00371},
	url          = {https://doi.org/10.1109/ICCV51070.2023.00371},
	bibsource    = {dblp computer science bibliography, https://dblp.org}
}

@article{wang_data-driven_2023,
	title        = {Data-driven methods for detection of abnormal ship behavior: Progress and trends},
	author       = {Yukuan Wang and Jingxian Liu and Ryan Wen Liu and Yang Liu and Zhi Yuan},
	year         = 2023,
	journal      = {Ocean Engineering},
	volume       = 271,
	pages        = 113673,
	doi          = {https://doi.org/10.1016/j.oceaneng.2023.113673},
	issn         = {0029-8018},
	url          = {https://www.sciencedirect.com/science/article/pii/S0029801823000574}
}

@phdthesis{sorensen_maritime_2024,
	title        = {Maritime Surveillance Finding Dark Ships with Satellites and Artificial Intelligence},
	author       = {Sørensen, Kristian Aalling},
	year         = 2024,
	publisher    = {Technical University of Denmark},
	language     = {English},
	school       = {Technical University of Denmark}
}

@inproceedings{li_ship_2017,
	title        = {Ship detection in SAR images based on an improved faster R-CNN},
	author       = {Li, Jianwei and Qu, Changwen and Shao, Jiaqi},
	year         = 2017,
	booktitle    = {2017 SAR in Big Data Era: Models, Methods and Applications (BIGSARDATA)},
	volume       = {},
	number       = {},
	pages        = {1--6},
	doi          = {10.1109/BIGSARDATA.2017.8124934}
}

@article{wei_hrsid_2020,
	title        = {{HRSID:} {A} High-Resolution {SAR} Images Dataset for Ship Detection and Instance Segmentation},
	author       = {Shunjun Wei and Xiangfeng Zeng and Qizhe Qu and Mou Wang and Hao Su and Jun Shi},
	year         = 2020,
	journal      = {{IEEE} Access},
	volume       = 8,
	pages        = {120234--120254},
	doi          = {10.1109/ACCESS.2020.3005861},
	url          = {https://doi.org/10.1109/ACCESS.2020.3005861},
	bibsource    = {dblp computer science bibliography, https://dblp.org}
}

@article{kang_contextual_2017,
	title        = {Contextual Region-Based Convolutional Neural Network with Multilayer Fusion for {SAR} Ship Detection},
	author       = {Miao Kang and Kefeng Ji and Xiangguang Leng and Zhao Lin},
	year         = 2017,
	journal      = {Remote. Sens.},
	volume       = 9,
	number       = 8,
	pages        = 860,
	doi          = {10.3390/RS9080860},
	url          = {https://doi.org/10.3390/rs9080860},
	bibsource    = {dblp computer science bibliography, https://dblp.org}
}

@article{zhang_sar_2021,
	title        = {{SAR} Ship Detection Dataset {(SSDD):} Official Release and Comprehensive Data Analysis},
	author       = {Tianwen Zhang and Xiaoling Zhang and Jianwei Li and Xiaowo Xu and Baoyou Wang and Xu Zhan and Yanqin Xu and Xiao Ke and Tianjiao Zeng and Hao Su and Israr Ahmad and Dece Pan and Chang Liu and Yue Zhou and Jun Shi and Shunjun Wei},
	year         = 2021,
	journal      = {Remote. Sens.},
	volume       = 13,
	number       = 18,
	pages        = 3690,
	doi          = {10.3390/RS13183690},
	url          = {https://doi.org/10.3390/rs13183690},
	bibsource    = {dblp computer science bibliography, https://dblp.org}
}

@article{wang_automatic_2019,
	title        = {Automatic Ship Detection Based on RetinaNet Using Multi-Resolution Gaofen-3 Imagery},
	author       = {Yuanyuan Wang and Chao Wang and Hong Zhang and Yingbo Dong and Sisi Wei},
	year         = 2019,
	journal      = {Remote. Sens.},
	volume       = 11,
	number       = 5,
	pages        = 531,
	doi          = {10.3390/RS11050531},
	url          = {https://doi.org/10.3390/rs11050531},
	bibsource    = {dblp computer science bibliography, https://dblp.org}
}

@article{sun_bifa-yolo_2021,
	title        = {BiFA-YOLO: {A} Novel YOLO-Based Method for Arbitrary-Oriented Ship Detection in High-Resolution {SAR} Images},
	author       = {Zhongzhen Sun and Xiangguang Leng and Yu Lei and Boli Xiong and Kefeng Ji and Gangyao Kuang},
	year         = 2021,
	journal      = {Remote. Sens.},
	volume       = 13,
	number       = 21,
	pages        = 4209,
	doi          = {10.3390/RS13214209},
	url          = {https://doi.org/10.3390/rs13214209},
	bibsource    = {dblp computer science bibliography, https://dblp.org}
}

@article{zhang_ls-ssdd-v10_2020,
	title        = {LS-SSDD-v1.0: {A} Deep Learning Dataset Dedicated to Small Ship Detection from Large-Scale Sentinel-1 {SAR} Images},
	author       = {Tianwen Zhang and Xiaoling Zhang and Xiao Ke and Xu Zhan and Jun Shi and Shunjun Wei and Dece Pan and Jianwei Li and Hao Su and Yue Zhou and Durga Kumar},
	year         = 2020,
	journal      = {Remote. Sens.},
	volume       = 12,
	number       = 18,
	pages        = 2997,
	doi          = {10.3390/RS12182997},
	url          = {https://doi.org/10.3390/rs12182997},
	bibsource    = {dblp computer science bibliography, https://dblp.org}
}
\end{document}